\documentclass{article}

\usepackage[preprint]{corl_2026} % Uncomment for pre-prints (e.g., arxiv); This is like ``final'', but will remove the CORL footnote.

\usepackage[utf8]{inputenc} % allow utf-8 input
\usepackage[T1]{fontenc}    % use 8-bit T1 fonts
\usepackage{hyperref}       % hyperlinks
\usepackage{url}            % simple URL typesetting
\usepackage{booktabs}       % professional-quality tables
\usepackage{amsfonts}       % blackboard math symbols
\usepackage{nicefrac}       % compact symbols for 1/2, etc.
\usepackage{multicol}
\usepackage{microtype}      % microtypography

\usepackage{mathtools}
\usepackage{graphics}
\usepackage{subcaption}
\usepackage[pdftex]{graphicx}
\usepackage{wrapfig}
\usepackage{caption}
\usepackage{color}
\usepackage{dsfont}
\usepackage[]{mdframed}
\usepackage{algorithm}
\usepackage{algorithmic}
\usepackage{xparse}
\usepackage{amsmath, amssymb, amsthm, amsfonts}
\usepackage{mathtools}
\usepackage{booktabs}
\usepackage{adjustbox}
\usepackage{array}
\usepackage{rotating}
\usepackage{multirow}
\usepackage{placeins}
\usepackage{xspace}
\usepackage[capitalise]{cleveref}
\usepackage{gensymb}
\usepackage{caption}
\usepackage{markdown}
\newcommand{\VisualEncoder}{SigLIP\,2~\cite{siglipv2}\xspace}

\crefname{equation}{Eq.}{Eqs.}
\Crefname{equation}{Eq.}{Eqs.}
\crefname{figure}{Fig.}{Figs.}
\Crefname{figure}{Fig.}{Figs.}
\crefname{table}{Tab.}{Tabs.}
\Crefname{table}{Tab.}{Tabs.}
\crefname{section}{Sec.}{Secs.}
\Crefname{section}{Sec.}{Secs.}

\def \MethodName {Unified Motion-Action\xspace}
\def \MethodAbbr {UMA\xspace}

\definecolor{linkpink}{RGB}{237,0,140}

\title{Unified Motion-Action Modeling for \\ Heterogeneous Robot Learning}
\author{
    \textbf{
    Yunhao Cao$^*$
    \quad
    Shitong Liu$^*$
    \quad
    Chao Feng$^*$
    \quad
    Meryl Zhang
    } \\[1mm]
    \textbf{
    Xuanchen Lu
    \quad
    Andrew Owens
    \quad
    Kuan Fang
    } \\[1mm]
    Cornell University \\[4mm]
    {\href{https://uma-manipulation.github.io}
      {\textcolor{linkpink}{\texttt{https://uma-manipulation.github.io}}}
    }
}

\begin{document}
\maketitle
\def\thefootnote{*}\footnotetext{Equal contribution, correspondence to \texttt{yunhao@cs.cornell.edu}}
% \begin{center}
% \vspace{-5mm}
% Cornell University \\
% \url{https://uma-manipulation.github.io}
% \end{center}

%===============================================================================

\begin{figure*}[h]
    % \vspace{-7mm}
    \centering
    \includegraphics[width=0.98\textwidth]{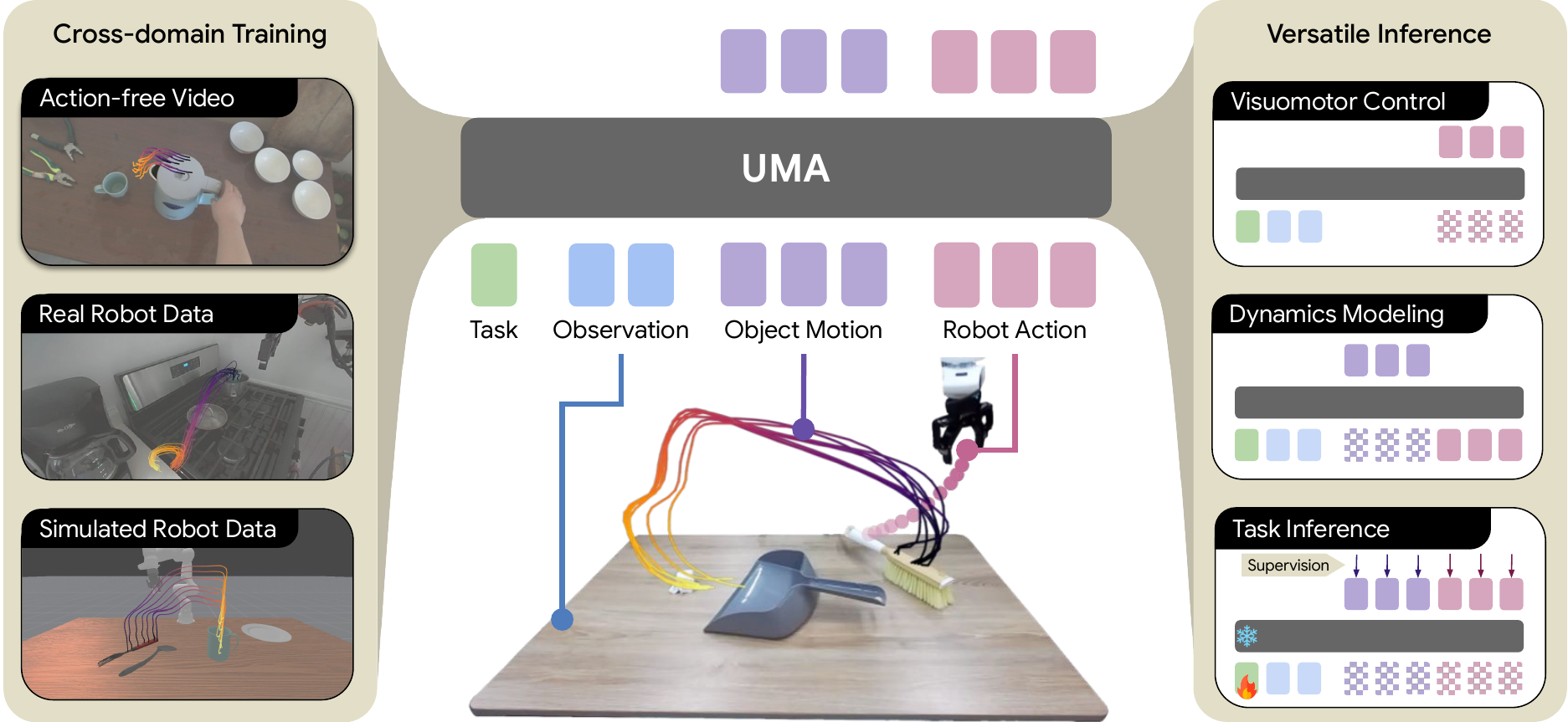}
    \caption{
    \textbf{Unified Motion-Action (UMA) Model.} 
    \MethodAbbr~uses object motion as a shared interface for heterogeneous robot learning. Pretraining effectively combines action-free videos, real robot data, and simulated robot data by representing task intent, observations, object motion, and robot actions as tokens under a masked generative objective. The same pretrained parameters then flexibly support visuomotor control, dynamics modeling, and task inference at deployment.
    }
    \vspace{3mm}
    \label{fig:teaser}
\end{figure*}
% \begin{figure*}[h]
%     \vspace{-5mm}
%     \centering
%     \includegraphics[width=1.0\textwidth]{figures/teaser.pdf}
%     \caption{
%         We introduce \MethodName~(\MethodAbbr)~Model, ...
%         \todo{Improve teaser and write caption}
%     }
%     \label{fig:teaser}
% \end{figure*}

\begin{abstract}
We present \MethodName~(\MethodAbbr)~Model, an approach that uses 3D object motion trajectories as a shared interface to bridge visuomotor control and dynamics modeling. \MethodAbbr~treats object motion and robot actions as co-evolving variables under a masked generative objective, in which the mask pattern determines both the supervision regime during pretraining and the inference mode at deployment. Using hindsight-relabeled motion contexts and a contrastive objective that disentangles task intent from scene geometry, \MethodAbbr~enables multi-task pretraining across heterogeneous data sources without requiring manually annotated task instructions. At deployment, the same pretrained parameters support motion-conditioned visuomotor control, motion-based dynamics modeling, and task adaptation from few-shot demonstrations. Pretrained on a mixture of robot demonstrations, human videos, and simulated data, \MethodAbbr~consistently outperforms state-of-the-art baselines specialized for each inference mode. 
% Videos and additional details are available at \url{https://uma-manipulation.github.io}.
\end{abstract}
% Building robotic foundation models that generalize across objects, scenes, tasks, and embodiments requires absorbing knowledge from heterogeneous data sources. Current vision--language--action and world-model approaches consume these sources through pixel-level or latent representations that entangle object motion with appearance, viewpoint, and embodiment, limiting cross-source transfer. We propose to address this bottleneck using \emph{dense 3D keypoint trajectories}: because trajectories describe how objects move independently of who is acting, they transfer across embodiments; and because they live in 3D space, they are directly comparable to robot actions without a lossy pixel-to-control mapping. We introduce \MethodName~(\MethodAbbr)~Model, a foundation model that jointly models keypoint trajectories and robot actions as co-evolving variables under a unified masked generative objective, enabling training on heterogeneous data without embodiment bias and flexible inference --- action generation, motion prediction, or motion-conditioned control --- by varying the conditioning mask at test time. Across diverse manipulation tasks, \MethodAbbr matches or exceeds state-of-the-art VLA baselines while using substantially less robot action data, improves as human video data is added, and adapts to new tasks from human demonstrations alone.
% \end{abstract}

%\keywords{CoRL, Robots, Learning} 
\vspace{2mm}
% \vspace{-5mm}

%===============================================================================

\newpage

\section{Introduction}
\label{sec:introduction}
\vspace{-5pt}

Learning from broad data and operating across diverse use cases is the defining promise of robotic foundation models.
Despite rapid progress in end-to-end and black-box approaches, including vision--language--action models~\cite{brohan_rt-2_2023, kim24openvla, pi0} and world models~\cite{hafner2020dreamerv2, yang2023unisim, agarwal2025cosmos}, they often fall short on transferring knowledge across domains and remain brittle outside the training distribution.
Effective scaling must therefore address not only the limits of robot data but also the absence of representations of physical interaction in which learning can be grounded.
Such representations should be structured enough to support sample-efficient learning, yet generic enough to support various settings.

Recently, object motion has emerged as a promising representation to ground visuomotor learning in physical interactions. When represented as 3D point trajectories on object surfaces tracked across time, object motion can be extracted from video without action labels, lives in the same spatial coordinate system as robot controls, and remains largely comparable across embodiments and camera views. Yet few existing methods exploit the full set of properties this representation affords. Motion-conditioned policies~\cite{xu2024im2flow2act, zhi_3dflowaction_2025} learn to predict robot actions from a given target motion but require paired motion-action labels, leaving action-free video data unused. Motion-prediction models~\cite{huang2026pointworld, yuan2024generalflow} forecast motion trajectories from video given language or visual goals but require a separate downstream controller and offer no straightforward way to incorporate action labels when robot data is available. As a result, no single existing method can both learn motion from action-free video and leverage action labels when they are available.

In this paper, we present \MethodName~(\MethodAbbr)~Model, an approach that uses object motion as a shared interface to bridge visuomotor control and dynamics modeling. As illustrated in \Cref{fig:teaser}, \MethodAbbr~treats object motion and robot actions as co-evolving variables, jointly predicted by a transformer-based sequence model from the current observation and task specification. Instead of fixing object motion in a single role, \MethodAbbr~allows both variables to serve as conditioning or as target depending on the mask pattern, letting heterogeneous data sources with and without action labels contribute supervision through one shared objective. We further extend hindsight experience replay~\cite{andrychowicz2017her} from goal-conditioned to motion-conditioned training, and combine it with a contrastive objective that yields a spatially invariant task representation. The same pretrained parameters then support action generation, dynamics prediction, and fast adaptation through soft prompt tuning at deployment. Unlike motion-conditioned policies from prior work~\cite{xu2024im2flow2act, coil2025}, our task representation is generic enough to perform new tasks even without manually specified reference motion.
We evaluate \MethodAbbr~on real-world tabletop manipulation spanning rigid objects, tool use, and deformable materials, pretraining on publicly available data with no overlap with the test environments. \MethodAbbr~ consistently outperforms the strongest specialized baseline in zero-shot and few-shot visuomotor control.

\section{Related Work}
\vspace{-5pt}

%chao: not a big deal but we don't have a keyword for paragraph in related work?

% \paragraph{Object Motion as a Representation for Robot Learning.}
Object motion, expressed as trajectories of keypoints or flows, has emerged as a structured intermediate representation, yet existing methods commit it to a single role per model. Motion-based methods work in either 2D image coordinates~\cite{wen2023atm, vecerik_robotap_2023, Gu2023RTTrajectoryRT, gao2024flip, xu2024im2flow2act, ren2025motion}, which require the controller to absorb a camera-to-3D mapping, or 3D scene coordinates~\cite{zhi_3dflowaction_2025, haldar_point_2025, coil2025, yuan2024generalflow, dharmarajan2025dream2flow, hsu_spot_2024}, which share a coordinate frame with robot actions. In parallel, particle- and point-based forward dynamics models predict scene response to candidate actions, from per-instance deformable models~\cite{li2018learningparticledynamics, zhang2024pgnd} to large-scale point-flow world models~\cite{huang2026pointworld}. Despite these advances, methods specialize in either action generation, which requires paired motion-action labels and cannot ingest action-free video, or dynamics modeling, which requires a separate controller and lacks a native interface for action labels. Our approach instead jointly models both within a single masked generative policy, supporting motion-conditioned action generation, action-conditioned motion generation, and soft-prompt task adaptation from the same parameters by varying the conditioning mask.

% \paragraph{Generalist Robot Models and Learned Dynamics.}
Our approach also relates to generalist policies that cast control as sequence modeling and world models that predict environment dynamics. Generalist policies include vision-language-action models~\cite{brohan_rt-2_2023, kim24openvla, octo_2023, pi0, pi05, liu2024rdt} and masked joint sequence models that unify dynamics and control~\cite{chen2021decisiontransformer, wu2023mpm, liu2022maskdp, radosavovic2023rpt, li2025uva}. These formulations either tie dynamics to embodiment-specific states~\cite{wu2023mpm, liu2022maskdp} or model them in raw pixels~\cite{radosavovic2023rpt, li2025uva, liu2024rdt, pi05}, leaving them respectively incompatible with cross-domain data or entangled with appearance and viewpoint. World models offer a complementary view, from latent rollouts for planning~\cite{ebert2018visualforesight, hafner2019planet, hafner2019dreamer, hafner2020dreamerv2} to video generators~\cite{yang2023unisim, agarwal2025cosmos, liao2025genieenvisioner, jang2025dreamgen, ali2025world, black2023susie}, predictive video features~\cite{bardes2024vjepa, assran2025vjepa2}, and physics-grounded digital twins~\cite{jiang2025phystwin}, but they operate in pixel or latent visual spaces that are expensive to roll out and provide no structured planning signal. Closer to our setting, concurrent works couple action and video generation within a single model~\cite{zhu2025unified, cen2025worldvla, zhang2025dreamvla}, yet still represent dynamics in pixel or token space without an embodiment-agnostic intermediate for action-free human video. Our approach grounds the joint formulation in object-centric 3D motion, yielding rollouts that share a coordinate frame with robot actions and natively absorb action-free human video. We further extend hindsight experience replay~\cite{andrychowicz2017her} from single-state goals to motion contexts, enabling multi-task pretraining without manual task labels.

\begin{figure*}[t]
    \centering
    \includegraphics[width=1.0\textwidth]{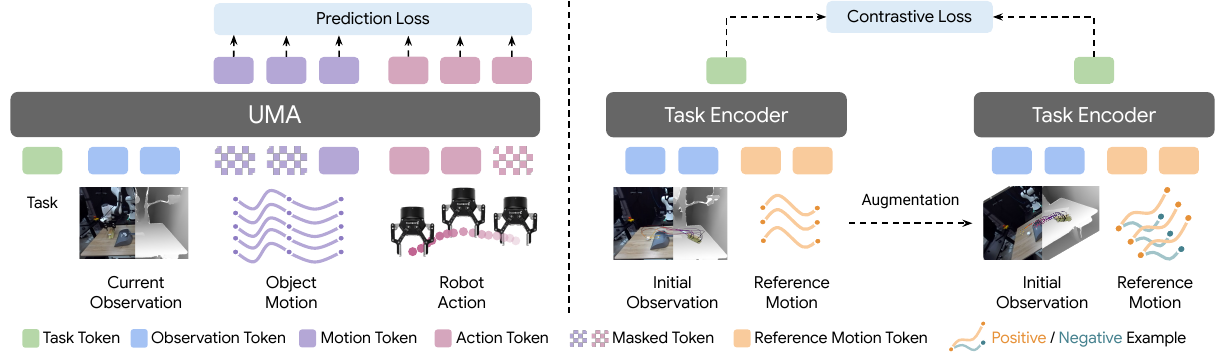}
    \vspace{-5mm}
    \caption{\textbf{Pre-Training of \MethodAbbr.}
        Left: \MethodAbbr is trained with a flow matching objective to predict randomly masked object motion and robot actions, conditioned on a task latent and visual observation.
        Right: We encode the reference motion and initial observation into task tokens, using both flow-matching and contrastive objectives to ensure semantic consistency of the learned task representation.
    }
    \vspace{-5mm}
    \label{fig:method_overview}
\end{figure*}

\section{\MethodName~Model}
\label{sec:method}
\vspace{-5pt}

We present \MethodName~(\MethodAbbr)~Model, which treats object motion and robot actions as co-evolving variables under a masked generative objective. Object motion serves as a structured yet generic intermediate between dynamics modeling and visuomotor control, enabling joint pretraining across heterogeneous data sources and supporting multiple inference modes.

% \vspace{}

\subsection{Bridging Control and Dynamics with Object Motion}
\label{subsec:object_motion_interface}
\vspace{-5pt}

We train a single model on heterogeneous data sources including action-free human videos, real-world robot data, and simulated robot data. Following prior work on sequence models for robotic control~\citep{li2025uva, chen2021decisiontransformer, liu2022maskdp}, the model jointly reconstructs object motion and action tokens under a shared masked flow-matching objective. Unlike prior joint-sequence formulations, we represent dynamics using object motion rather than raw observations. Since the motion stream is defined independently of any specific embodiment, data sources without compatible action labels can still provide supervision for dynamics, while datasets with action labels supervise both the dynamics and action streams.

Concretely, extending the definitions in prior work~\citep{xu2024im2flow2act, yuan2024generalflow, coil2025}, we represent object motion through 3D points sampled on object surfaces and tracked across time. Each sample is a 3D position $x_{k,t} \in \mathbb{R}^3$, where $k \in \{1, \ldots, K\}$ indexes keypoints and $t \in \{0, \ldots, T\}$ indexes timesteps, expressed in the camera coordinate frame. These trajectories can be extracted from monocular video via off-the-shelf trackers for training-time supervision, while at deployment \MethodAbbr~generates them implicitly through the same generative process that produces actions.

\textbf{Motion-action modeling.}
Let $o_t$ denote the current observation, $x_{t:t+H}$ the object motion sequence, and $a_{t:t+H}$ the action sequence over a prediction horizon $H$. We model the conditional distribution
\begin{equation}
    p_\theta\!\left(\,x_{\mathcal{P}_x},\; a_{\mathcal{P}_a}\;\big|\;o_t,\; c,\; x_{\mathcal{G}_x},\; a_{\mathcal{G}_a}\,\right),
    \label{eq:masked_joint_model}
\end{equation}
where $\mathcal{G}_x, \mathcal{G}_a$ index given conditioning tokens, $\mathcal{P}_x, \mathcal{P}_a$ index predicted tokens, and $c$ is a latent variable we call the \emph{task latent}, summarizing the abstract intent of the interaction in a form invariant to the specific scene and action space, and agnostic to the modality through which the intent is specified. Different choices of $(\mathcal{P}, \mathcal{G})$ instantiate the regimes used throughout the paper. During pretraining (\Cref{subsec:method_pretraining}), we sample random masks over both motion and action streams. For motion-conditioned action generation, the model conditions on $c$ derived from a reference motion and sets $\mathcal{P}_a$ to future actions. For action-conditioned dynamics prediction, used inside model predictive control, the model conditions on candidate future actions in $\mathcal{G}_a$ and sets $\mathcal{P}_x$ to the resulting object-motion rollout. For soft prompt tuning, the task latent $c$ is optimized while model weights remain fixed.

\textbf{Task inference from reference motion.}
A natural way to specify the task at training time is through a partial motion observation drawn from a recorded interaction of the target task. We call such a partial observation a \emph{reference motion},
\begin{equation}
x_c = \{\tilde{x}_{k,t} : (k,t) \in \mathcal{G}_c\},
    \quad \mathcal{G}_c \subset \{1,\ldots,K\} \times \{0,\ldots,T\},
    \label{eq:reference_motion}
\end{equation}
which may contain \textit{different keypoints}, \textit{different numbers of samples}, and \textit{nonuniform temporal spacing} across episodes. A reference motion cannot serve as $c$ directly because how the objects move from a particular initial scene is observation-dependent. Recovering an observation-agnostic $c$ therefore requires both the reference motion and the initial observation. We introduce a task encoder $h_\psi$ that consumes both and produces a fixed-size summary of task intent,
\begin{equation}
    c = h_\psi(o_0, x_c).
    \label{eq:task_latent_inference}
\end{equation}
By design, $h_\psi$ must produce a $c$ that is invariant to which keypoints and timesteps the reference motion samples and to the absolute scene configuration.

\subsection{Motion-Action Diffusion Transformer}
\label{subsec:method_architecture}
\vspace{-5pt}

The masked joint model imposes three requirements that motivate our architecture. All modalities must share a token space, the encoder must consume sparse irregular reference motions, and attention must respect local object-action coupling within timesteps and coherent evolution across time. \MethodAbbr~addresses these with a tokenization scheme, a task encoder $h_\psi$, and a masked diffusion transformer~\citep{Peebles2022DiT, dasari2024ditpi} with structured spatiotemporal attention~\citep{coil2025} that jointly denoises masked motion and action tokens.

\textbf{Tokenization.}
\MethodAbbr~represents observations, object motion, and robot actions as typed tokens sharing the transformer backbone. Observation tokens summarize the current RGB-D observation $o_t$. Each object-motion sample $x_{k,t}$ becomes one token projected through a lightweight MLP with keypoint-identity and time embeddings from its $(k, t)$ indices. Action tokens encode robot controls at corresponding timesteps via a separate input projection. The task latent $c$ is realized as a small set of task tokens inserted alongside the other modalities. Unlike VLA models~\citep{kim24openvla, pi0, pi05, liu2024rdt} that tokenize only pixels and actions, this scheme gives object motion, executable control, and task intent a common latent space while preserving modality-specific input geometry.

\textbf{Task encoder.}
The task encoder $h_\psi(o_0, x_c)$ produces $M$ task tokens comprising the task latent $c$ from a sparse reference motion and the initial observation. The initial observation $o_0$ is encoded into geometry-aware point-cloud features via a PointNet++~\citep{qi2017pointnet++} backbone consuming 3D positions concatenated with \VisualEncoder~image features. Since the reference motion specifies only waypoint ordering rather than absolute timesteps, each sample $x^c_{k,t}$ is projected through a lightweight MLP with time indices normalized to $[0,1]$, making $h_\psi$ invariant to the total number of timesteps and to irregular temporal spacing. These tokens are cross-attended to the point-cloud features to produce $c$.

\textbf{Joint denoising with structured attention.}
Object motion and robot actions are two coupled views of the same physical interaction. Modeling only actions, as in standard diffusion policies~\citep{chi2025diffusion}, cannot exploit action-free videos, while modeling only motion needs a separate controller to translate predicted dynamics into executable actions. \MethodAbbr~jointly predicts both via a shared denoising backbone trained with flow matching~\citep{lipman_flow_matching_2023}, so action-labeled robot data ties motion to control while action-free video data still supervises the motion stream. The observation, task latent $c$, and unmasked tokens $(x_{\mathcal{G}_x}, a_{\mathcal{G}_a})$ remain clean as conditioning, while masked targets are iteratively denoised, with separate final layers preserving the different output geometries of 3D keypoint positions and robot controls. Following COIL~\citep{coil2025}, attention is factorized into \emph{spatial} (within-timestep motion-action coupling), \emph{temporal} (same-point across timesteps), and \emph{context} (target tokens to observations) patterns. Unlike COIL, which applies these inside standard transformer layers, \MethodAbbr~implements them within DiT blocks~\citep{Peebles2022DiT, dasari2024ditpi} that condition on diffusion time via adaptive layer norm. By assigning separate modulation weights to masked and unmasked tokens, these \emph{Masked DiT blocks} apply stronger denoising to targets while preserving unmodulated conditioning, without dense global attention.

\subsection{Cross-Domain Training via Masked Autoencoding}
\label{subsec:method_pretraining}
\vspace{-5pt}

Pretraining instantiates \Cref{eq:masked_joint_model} as masked autoencoding over object-motion and action tokens. For each episode, we sample a reference motion $x_c$, encode it via $h_\psi(o_0, x_c)$ into a task latent $c$, and sample masks $(\mathcal{G}_x, \mathcal{G}_a)$ across modality, time, and space to determine which tokens are reconstructed. The joint denoiser and task encoder are trained together, with only available target tokens contributing to the loss. This generalizes hindsight goal relabeling, where the hindsight ``goal'' is a partial spatiotemporal motion specification rather than a single state.

\textbf{Heterogeneous supervision.}
We write the per-modality reconstruction losses as $\mathcal{L}_x$ and $\mathcal{L}_a$, both flow-matching losses~\citep{lipman_flow_matching_2023} on the noised target tokens (details in Appendix). The mask-based formulation lets each source supervise only the terms for which it has labels. Target-robot demonstrations supervise both $\mathcal{L}_x$ and $\mathcal{L}_a$, while action-free sources such as human videos and cross-embodiment data with incompatible actions supervise only $\mathcal{L}_x$. All sources represent motion, geometry, and actions in the local camera frame of each episode, and predicted actions are transformed to the robot base frame at deployment via calibrated extrinsics. Action-free datasets require no calibration since they supervise only motion tokens.

\textbf{Ensuring task latent consistency.}
For $c$ to support transfer across observations and adaptation to new tasks, it must summarize task intent without carrying spurious details of the specific reference motion or absolute scene placement. We promote this disentanglement with a SimCLR-style contrastive objective $\mathcal{L}_c$~\citep{chen2020simple} over augmented views of the same $(o_0, x_c)$ pair. Positive views are produced by keypoint subsampling, timestep subsampling, and random SE(3) transformations applied jointly to $o_0$ and $x_c$ at the encoder input, while the diffusion transformer tokens remain in the original camera frame. Negative views come from different tasks or different object-motion intents. Because $c$ carries no information about which SE(3) transformation was applied at the encoder, the decoder cannot exploit absolute pose through $c$ and is forced to interpret it as pose-invariant task intent.

\textbf{Full pretraining objective.}
The full pretraining loss combines motion reconstruction, action reconstruction, and task-latent disentanglement,
\begin{equation}
    \mathcal{L}
    =
    \lambda_x\, \mathcal{L}_x
    +
    \lambda_a\, \mathcal{L}_a
    +
    \lambda_c\, \mathcal{L}_c.
    \label{eq:training_loss}
\end{equation}
The three terms are optimized jointly across the heterogeneous data mixture. The mask-based supervision naturally accommodates missing modalities, while $\mathcal{L}_c$ acts on the task encoder regardless of whether the source provides action labels, so action-free data contributes to both $\mathcal{L}_x$ and $\mathcal{L}_c$.

\subsection{Versatile Inference}
\label{subsec:method_inference}

At test time, \MethodAbbr~changes which tokens are given and which are predicted while keeping the model weights fixed, except in soft prompt tuning. This gives three inference modes.

\textbf{Motion-conditioned visuomotor control.}
Given the initial observation $o_0$ of the deployment scene and a target reference motion $x^\star$, the task encoder produces $c = h_\psi(o_0, x^\star)$, and \MethodAbbr~predicts future actions from the current observation $o_t$,
\begin{equation}
    \hat{a}_{t:t+H}
    \sim
    p_\theta
    \left(
        a_{t:t+H}\;\big|\;o_t,\; c,\; x_{\mathcal{G}_x},\; a_{\mathcal{G}_a}
    \right),
    \label{eq:motion_conditioned_policy}
\end{equation}
where the given motion and action tokens may include recent history. This converts a target object motion into executable robot actions. At each step, the model samples an action chunk, transforms it to the robot base frame, executes, and replans from the new observation, letting the policy correct deviations between target and realized object motion in closed loop. Because $c$ captures task intent in a disentangled latent space, $p_\theta$ can be conditioned on alternative task representations such as language instructions without retraining the policy. See Appendix for details.

\textbf{Motion-based dynamics modeling.}
\MethodAbbr~can also be used in the forward direction as a dynamics model. Given a candidate action sequence $a^{(j)}_{t:t+H}$, the model predicts the resulting object motion,
\begin{equation}
    \hat{x}^{(j)}_{t:t+H}
    \sim
    p_\theta
    \left(
        x_{t:t+H}\;\big|\;o_t,\; c,\; x_{\mathcal{G}_x},\; a^{(j)}_{t:t+H}
    \right).
    \label{eq:motion_prediction}
\end{equation}
This supports model predictive control by scoring candidate actions against the target reference motion, and additionally enables diagnostic interpretability by exposing whether predicted dynamics match task intent before any action is executed.

\textbf{Task inference via soft prompt tuning.}
For a new task, \MethodAbbr~can adapt by treating $c$ as a free optimization variable rather than the output of $h_\psi$, keeping all other parameters frozen. Given a small demonstration set $\mathcal{D}$, we initialize $c$ at $c_0$ and optimize
\begin{equation}
    c^\star
    =
    \arg\min_{c}
    \sum_{\xi \in \mathcal{D}}
    \left[
        \lambda'_x\,
        \mathcal{L}_{x}(\xi;\, c)
        +
        \lambda'_a\,
        \mathcal{L}_{a}(\xi;\, c)
    \right]
    +
    \lambda'_{\mathrm{c}}\, \|c - c_0\|_2^2,
    \label{eq:prompt_adaptation}
\end{equation}
with $\lambda'_x, \lambda'_a > 0$ for action-supervised demonstrations and $\lambda'_a = 0$ for video-only demonstrations, so the latter contribute only motion supervision. This setting is robot-free for the new task, relying on the pretrained model's executable-action knowledge from prior robot data and on the disentanglement of $c$ established above to transfer the optimized prompt across scenes.

\section{Experiments}
\vspace{-5pt}

Through experiments and analysis, we investigate \textbf{Q1:} whether a single pre-trained \MethodAbbr~Model
matches or exceeds state-of-the-art baselines specialized for each
use case, \textbf{Q2:} how heterogeneous data sources contribute to
performance, and \textbf{Q3:} which design choices in \MethodAbbr~are
essential.

\subsection{Experimental Setup}
\label{subsec:experiment-setup}
\vspace{-5pt}
    
We evaluate \MethodAbbr~on real-world tabletop manipulation following the DROID setting~\cite{khazatsky2024droid}. Pre-training uses a mixture of DROID demonstrations, human videos from HOI4D~\cite{Liu_2022_hoi4d} and Xperience~\cite{xperience_10m}, and simulated robot data with randomized objects and motions, none of which overlaps the real-world test environments. As shown in \Cref{fig:zero_shot_quantitative}, the real-world evaluation covers 6-DoF manipulation (\textit{Insertion}), tool use (\textit{Sweeping}), and deformable object manipulation (\textit{Folding}), with three additional simulated tasks used for ablations. Each task is evaluated over 20 trials with randomized scenes and camera configurations. Further details on the tasks and the data pipelines can be found in the Appendix.

% ------------------------------------------------------------------
\subsection{Zero-Shot Generalization}
\label{subsec:experiments_zero_shot}
\vspace{-5pt}

To answer \textbf{Q1} in the zero-shot setting, we test \MethodAbbr on motion-conditioned visuomotor control and dynamics prediction by changing only which tokens are given and which are predicted in \Cref{eq:masked_joint_model}.
\vspace{-5pt}

\begin{figure*}[t]
    \centering
    \begin{subfigure}[t]{0.49\textwidth}
        \centering
        \includegraphics[width=\linewidth]{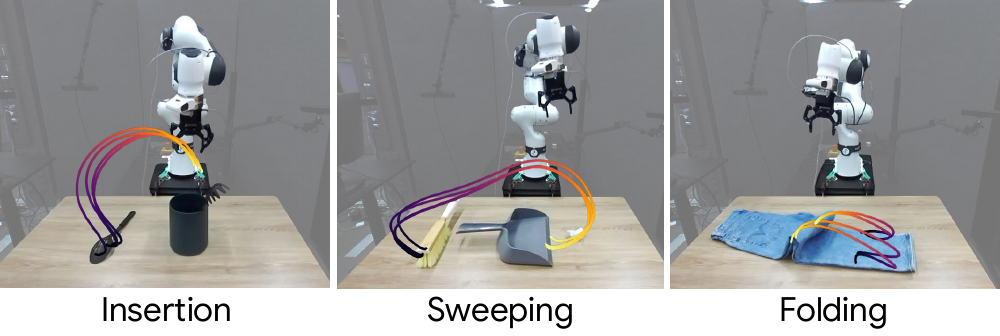}
        % \caption{Real-world evaluation tasks.}
    \end{subfigure}
    \hfill
    \begin{subfigure}[t]{0.5\textwidth}
        \centering
        \includegraphics[width=\linewidth]{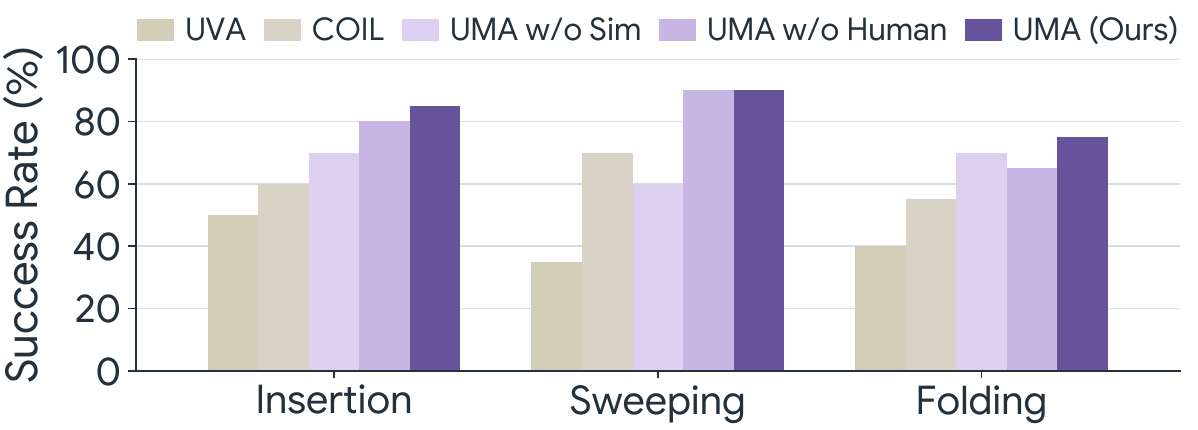}
        % \caption{Zero-shot visuomotor control.}
    \end{subfigure}
    \vspace{-3mm}
    \caption{
        \textbf{Zero-shot evaluation.} Left: real-world evaluation tasks used throughout our experiments. Right: success rates for motion-conditioned visuomotor control without task-specific finetuning.
    }
    \vspace{-3mm}
    \label{fig:task_qualitative}
    \label{fig:zero_shot_action}
    \label{fig:zero_shot_quantitative}
\end{figure*}

\paragraph{Visuomotor control.} For this inference mode (\Cref{eq:motion_conditioned_policy}), we compare against two strong baseline policies, both trained on the same data mixture as \MethodAbbr. \textit{COIL}~\cite{coil2025} is a 3D motion-conditioned flow-matching policy that does not model motion as a prediction target. \textit{UVA}~\cite{li2025uva} is a joint video-action masked model on raw pixels. \MethodAbbr~outperforms the strongest baseline by 20 to 25 percentage points on every task (\Cref{fig:zero_shot_action}). The improvement over COIL~\cite{coil2025} suggests that action prediction benefits from jointly modeling motion and action as coupled variables, rather than using motion only as a conditioning signal. 
The gap to UVA~\cite{li2025uva} further suggests that joint modeling is more effective when the dynamics are represented as motion rather than raw pixels, reducing appearance variations and aligning more naturally with robot actions.

\vspace{-5pt}

\begin{wraptable}{r}{0.32\textwidth}
    % \vspace{-1.0em}
    \centering
    \small
    \setlength{\tabcolsep}{8pt}
    \renewcommand{\arraystretch}{1.1}
    \begin{tabular}{lc}
        \toprule
        \textbf{Method} & \textbf{MSE} $\downarrow$ \\
        \midrule
        PointWorld~\cite{huang2026pointworld} & 0.054 \\
        \MethodAbbr~w/o Sim & 0.208 \\
        \MethodAbbr~w/o Human & 0.044 \\
        \textbf{\MethodAbbr~(Ours)} & \textbf{0.042} \\
        \bottomrule
    \end{tabular}
    \caption{\textbf{Dynamics prediction.} Mean squared error (MSE) evaluated across real-world tasks.}
    \label{tab:zero_shot_motion}
    \vspace{-1.3em}
\end{wraptable}

\paragraph{Dynamics prediction.} For dynamics prediction (\Cref{eq:motion_prediction}), we compare against \textbf{PointWorld}~\cite{huang2026pointworld}, a state-of-the-art action-conditioned keypoint dynamics model. \MethodAbbr~reduces motion-prediction MSE (in squared meters) from 0.054 to 0.042 (\Cref{tab:zero_shot_motion}), showing that joint motion-action training produces sharper dynamics than a dedicated dynamics model even when only motion is queried at test time. Notably, removing simulated data causes a 5x increase in MSE, indicating that randomized actions and states in simulation regularize the model against out-of-distribution configurations encountered in the real world.

% ----------------------------------------------------------------------

\subsection{Few-Shot Adaptation}
\label{subsec:experiments_few_shot}

For \textbf{Q1} in the few-shot setting, \MethodAbbr~adapts to a new task by optimizing only the task latent $c$ on 25 demonstrations while keeping all other parameters frozen (\Cref{eq:prompt_adaptation}). We consider two supervision regimes, robot demonstrations with action labels and human videos with motion-only supervision.

\vspace{-5pt}

\paragraph{Action supervision.}
Both the motion- and action-reconstruction terms in \Cref{eq:prompt_adaptation} are active ($\lambda'_x, \lambda'_a > 0$). We compare against two baselines that update model weights rather than a task latent: \textit{$\pi_{0.5}$} finetuned with LoRA~\cite{pi05, hu2022lora}, and \textit{UVA}~\cite{li2025uva} finetuned with action supervision. Despite updating only a low-dimensional task latent rather than millions of parameters, \MethodAbbr~matches the strongest baseline on Insertion and outperforms it by 10-25\% on Sweeping and Folding (\Cref{fig:few_shot_adaptation}).

\vspace{-5pt}

\paragraph{Motion supervision.}
Only the motion-reconstruction term is active ($\lambda'_a = 0$). In this setting, we record human videos performing the same task in the evaluation scene and extract the reference motion as well as motion tokens to provide supervision for adaptation. This rules out $\pi_{0.5}$, whose adaptation requires action labels. We instead finetune \textit{UVA}~\cite{li2025uva} with a pixel-reconstruction loss as the baseline. \MethodAbbr~outperforms UVA by approximately 25 percentage points on every task, showing that joint motion-action pretraining provides a more effective bridge from action-free human videos to robot control than pixel-action joint pretraining.

% ----------------------------------------------------------------------

\subsection{Cross-Domain Training Analysis}
\label{subsec:heterogeneous_data}
\vspace{-5pt}

To address \textbf{Q2}, we compare two data-ablated variants against the full model, as shown in \Cref{fig:zero_shot_quantitative} and \Cref{tab:zero_shot_motion}. Removing simulation (\textit{w/o Sim}) increases motion-prediction error by nearly 5$\times$ (MSE 0.042 to 0.208) and reduces zero-shot action success by up to 30 percentage points, confirming that dense simulated action-motion pairs are the primary source of motion-action coupling. Removing human videos (\textit{w/o Human}) leaves motion prediction nearly unchanged (0.042 to 0.044) but reduces action success especially on Folding (75\% to 65\%), suggesting that action-free human videos contribute task-level diversity that aids real-world transfer rather than dynamics accuracy. The two sources thus play complementary roles, with simulation strengthening motion-action coupling and human videos broadening task coverage.

% ----------------------------------------------------------------------

\subsection{Ablation and Failure Analysis}
\label{subsec:ablation_study}

\vspace{-5pt}

\begin{figure}[t]
    \vspace{-1.5em}
    \centering

    \begin{minipage}[t]{0.60\textwidth}
        \centering
        \vspace{0pt}

        \includegraphics[width=\linewidth]{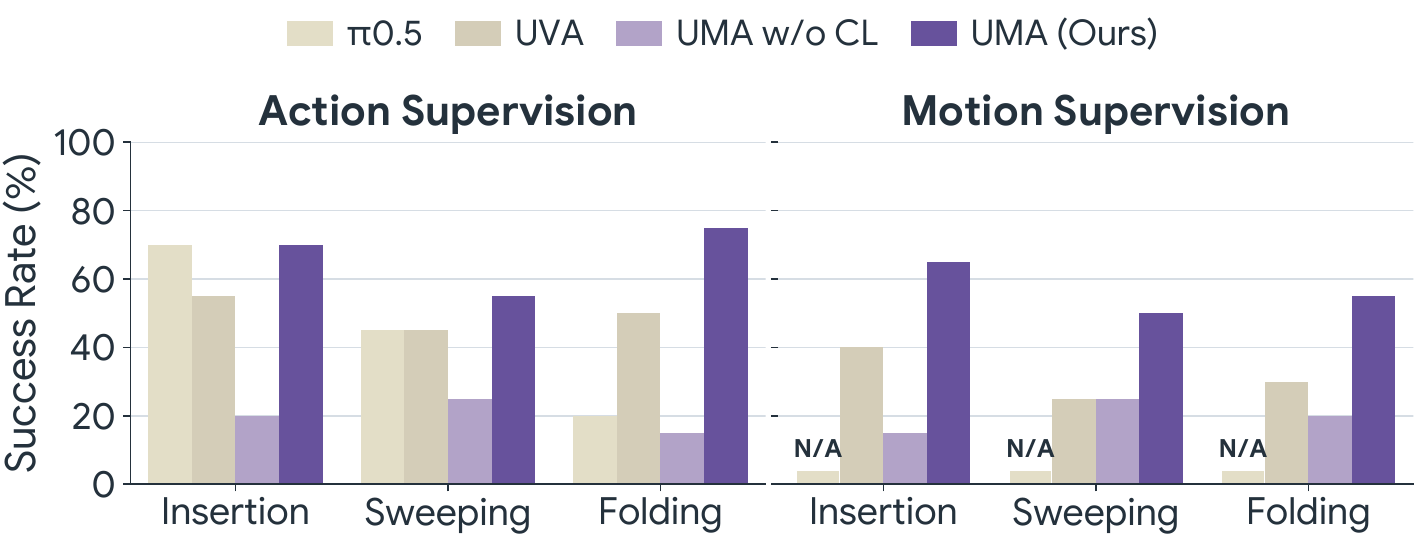}

        \vspace{-0.2em}
        \captionof{figure}{\textbf{Few-shot adaptation.}
        Success rates for adapting to new tasks from 25 target demonstrations under action supervision and motion supervision. 
        }
        \label{fig:few_shot_adaptation}
    \end{minipage}
    \hfill
    \begin{minipage}[t]{0.37\textwidth}
        \centering
        \vspace{0pt}
        \setlength{\fboxsep}{4pt}
        % \fbox{
        % \begin{minipage}[c][0.12\textheight][c]{0.88\linewidth}
        %     \centering
        %     \textbf{Failure Analysis Placeholder}\\[0.5em]
        %     \small Grasping \quad | \quad Execution
        % \end{minipage}
        % }
        \includegraphics[width=\linewidth]{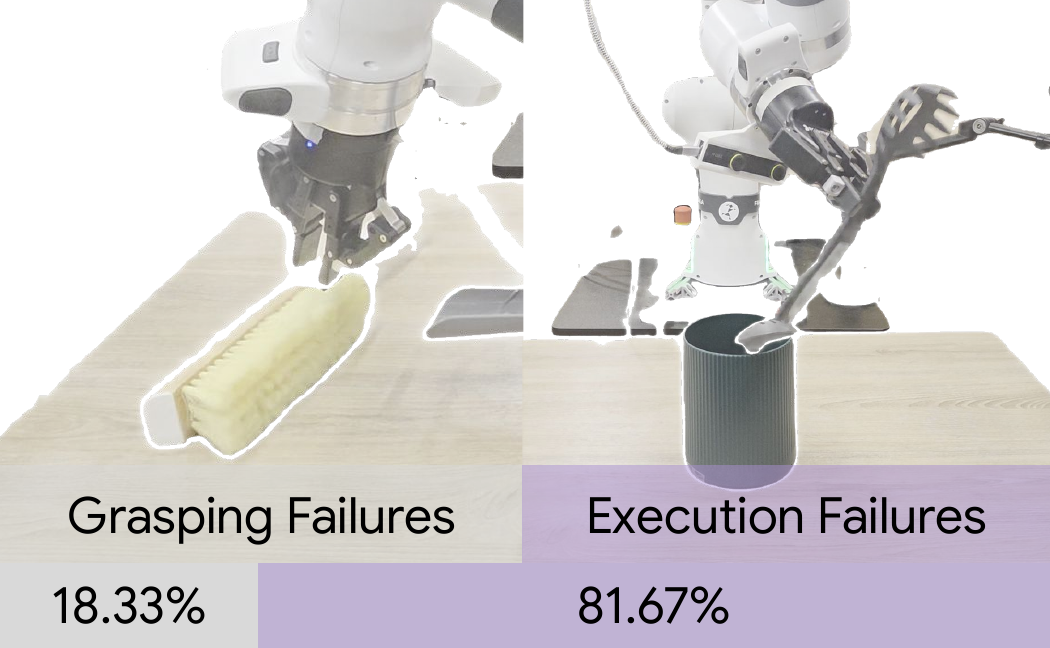}
        \vspace{-1.0em}
        \captionof{figure}{\textbf{Failure analysis.}
        Breakdown of failure cases by error type.}
        \label{fig:failure_analysis}
    \end{minipage}
    \vspace{-5mm}
\end{figure}

\paragraph{Effects of contrastive learning.}
The contrastive objective $\mathcal{L}_c$ trains the task latent $c$ to capture only the abstract intent of the task, remaining invariant to randomization of the initial scene. Removing the contrastive objective (\textit{w/o CL}) reduces few-shot success rates by 30 to 60 percentage points across both supervision regimes (\Cref{fig:few_shot_adaptation}), confirming that the contrastive grouping is what produces a transferable task latent. Without it, $c$ captures scene-specific trajectory details rather than transferable task intent, and few-shot adaptation reduces to overfitting on the specific reference motions.

\vspace{-5pt}

\begin{wrapfigure}{r}{0.28\textwidth}
    \vspace{-1.6em}
    \centering
    \includegraphics[width=\linewidth]{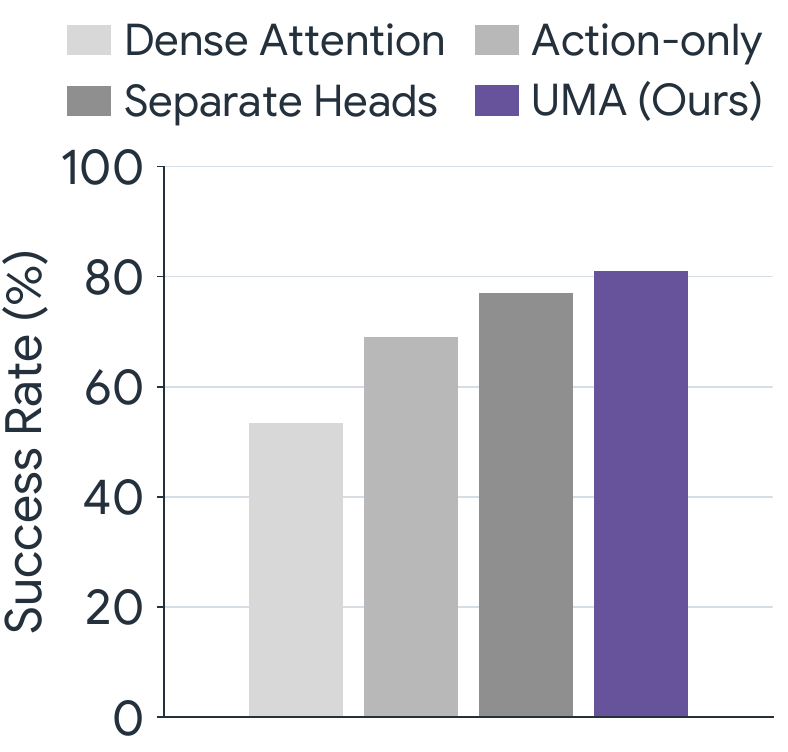}
    \vspace{-6mm}
    \caption{\textbf{Ablation study.}
    We evaluate the average success rates over simulation.}
    \label{fig:ablation}
    \vspace{-1.2em}
\end{wrapfigure}
\paragraph{Model design ablation.}
To address \textbf{Q3}, we evaluate three architecture variants trained on simulated robot data across three simulated tasks of 100 episodes each (\Cref{fig:ablation}): \textit{Action-only Supervision} (no motion prediction), \textit{Separate Heads} (independent decoder heads), and \textit{Dense Attention} (full dense attention at matched parameter count and training budget). Jointly predicting motion alongside actions is essential, with \textit{Action-only} trailing \MethodAbbr~by a wide margin especially on tool use, where coupled object-action dynamics dominate. Separate heads improves over action-only but remains slightly worse on average, indicating that a shared head better captures motion-action coupling. Dense attention underperforms at matched budget, showing that the structured attention facilitates information fusion.

\vspace{-5pt}

\paragraph{Failure analysis.}
Execution failures dominate at 82\% of total errors, with grasping accounting for the remaining 18\% (\Cref{fig:failure_analysis}). \MethodAbbr~thus reliably identifies the correct task intent and grasp configuration, but loses precision during the action rollout. This aligns with our finding in \Cref{subsec:heterogeneous_data} that simulation drives motion-action coupling, pointing toward further scaling of motion-action paired data, rather than richer task specification, as the most direct path to closing the remaining gap.

\section{Limitations}
\vspace{-5pt}

The current instantiation of \MethodAbbr~faces several limitations that suggest directions for future work. First, although \MethodAbbr~ingests heterogeneous data through the shared object-motion representation, deploying actions on a new embodiment still requires supervising an embodiment-specific action head from compatible robot data. Second, all representations live in a per-episode camera frame, so deploying robot actions requires calibrated camera-to-base extrinsics. Third, the current design conditions on a single visual observation per prediction step, leaving multi-step observation history and dense cross-frame point correspondences unexploited as supervision signals that fit naturally into the masked joint formulation.

\section{Conclusion}
\vspace{-5pt}

We presented \MethodName, a robot foundation model that jointly models 3D object motion and robot actions through a masked diffusion transformer, enabling heterogeneous pretraining from action-free human videos, simulated data, and real robot demonstrations. A single pretrained checkpoint supports zero-shot motion-conditioned control, action-conditioned dynamics prediction, and few-shot adaptation to new tasks via soft prompt tuning from either robot demonstrations or human videos. Our results suggest that object motion provides an effective shared representation for bridging dynamics modeling and visuomotor control, and that scaling motion pretraining is a promising path toward more generalizable manipulation.

\newpage
\acknowledgments{
% \textbf{Acknowledgments.}
We gratefully acknowledge use of the research computing resources of the Empire AI Consortium, Inc~\cite{Bloom2025EmpireAI}, with support from Empire State Development of the State of New York, the Simons Foundation, and the Secunda Family Foundation.
This work was supported in part by the Amazon Research Awards, an NVIDIA Academic Grant, and NSF CAREER \#2339071. We thank Chuanruo Ning, Tianrui Wang, Cory Fan, Xingyi He, Qianxu Wang, Qi Wu, and Pranav Thakkar for their constructive feedback.
}

% \newpage
\bibliography{references}  % .bib

\newpage

\appendix

\section{Implementation Details}

This section provides implementation details that complement the architectural overview in the main paper. We describe the contrastive learning objective used to train the task encoder, including the augmentation strategy and hard negative construction, followed by a detailed specification of the Masked DiT blocks that enable dual-role token processing within a single transformer pass, and finally the hyperparameters used in training the full UMA model.

\subsection{Contrastive Learning Objective}
\label{sec:contrastive_objective}

The contrastive loss $\mathcal{L}_c$ encourages the task encoder $h_\psi$ to produce task latents that are invariant to input augmentations while remaining discriminative across distinct manipulation intents. For each sample in a minibatch, we construct multiple views of the same task specification $(o_0, x_c)$ by independently applying keypoint subsampling, timestep subsampling, and random SE(3) transformations jointly to the initial point cloud $o_0$ and reference motion $x_c$. One view is left unaugmented; its task latent $c$ is used both to condition the flow-matching decoder and to participate in the contrastive loss. The remaining augmented views are used only for contrastive learning and never condition the denoising transformer. Negatives are views from other samples in the minibatch. We also construct hard negatives within the same scene by randomly interpolating SE(3) waypoints from the same starting keypoints but with different motion trajectories, yielding task specifications that share scene geometry but encode different manipulation intents.

The same contrastive formulation can incorporate additional task-specification modalities when they are available during pretraining. For example, given a language description $\ell$ of the same interaction, a language encoder $h_\omega(\ell)$ produces language task tokens that are inserted into the same contrastive pool as the motion-derived task tokens. Views from different modalities but the same underlying task are treated as positives, while views from different tasks or different motion intents remain negatives. This cross-modal contrastive term provides additional regularizations on the shared task-latent space, but is not required by the motion-action objective and is only used when such auxiliary annotations are available.

Let $B$ denote the minibatch size and $S$ denote the number of task-specification views per sample. For sample $i \in [0,B)$ and view $s \in [0,S)$, let $v_{i,s}=(o_{0,i,s}, x_{c,i,s})$ denote the corresponding view. Each view is encoded by $h_\psi$ into $M$ task tokens $c_{i,s}=h_\psi(v_{i,s})$. We flatten these tokens and pass them through a two-layer projection head $g_\phi$, followed by $\ell_2$ normalization:
\begin{equation}
    z_{i,s}
    =
    \frac{
    g_\phi\!\left(\mathrm{vec}(c_{i,s})\right)
    }{
    \left\|
    g_\phi\!\left(\mathrm{vec}(c_{i,s})\right)
    \right\|_2
    }
\end{equation}
 We use a SimCLR-style InfoNCE objective with multiple positives per anchor. For anchor $(i,s)$, the positive set is $\mathcal{P}_{i,s}=\{(i,r):r\neq s\}$ and the comparison set is $\mathcal{Q}_{i,s}=\{(j,r):(j,r)\neq(i,s)\}$. The loss is
\begin{equation}
    \mathcal{L}_c
    =
    -\frac{1}{|\mathcal{A}|}
    \sum_{(i,s)\in\mathcal{A}}
    \log
    \frac{
    \sum_{(i,r)\in\mathcal{P}_{i,s}}
    \exp\!\left( z_{i,s}^{\top} z_{i,r} / \tau \right)
    }{
    \sum_{(j,r)\in\mathcal{Q}_{i,s}}
    \exp\!\left( z_{i,s}^{\top} z_{j,r} / \tau \right)
    },
\end{equation}
where $\mathcal{A}$ is the set of valid anchors with at least one positive view and $\tau=0.07$ is the temperature. Gradients from $\mathcal{L}_c$ update only the task encoder $h_\psi$ and projection head $g_\phi$; the denoising transformer is trained only through the flow-matching reconstruction losses.

\subsection{Masked DiT Blocks}

The main paper introduces Masked DiT blocks as the mechanism that reconciles two conflicting requirements within a single transformer pass. Target tokens, which correspond to masked motion or action variables, must be iteratively denoised from a noisy state toward the data manifold. Given tokens, which correspond to unmasked motion or action variables provided as conditioning, are already clean and should pass through the network with minimal corruption so that they serve as reliable context for the denoising process. A standard DiT block~\citep{Peebles2022DiT} applies a single set of adaptive layer norm (adaLN) modulation weights, conditioned on the diffusion timestep $t$, uniformly to all tokens. This creates a tension: modulation tuned for denoising (large scale shifts that depend on noise level) is inappropriate for clean tokens, and vice versa.

\begin{figure}[t]
    \centering
    \includegraphics[width=0.95\textwidth]{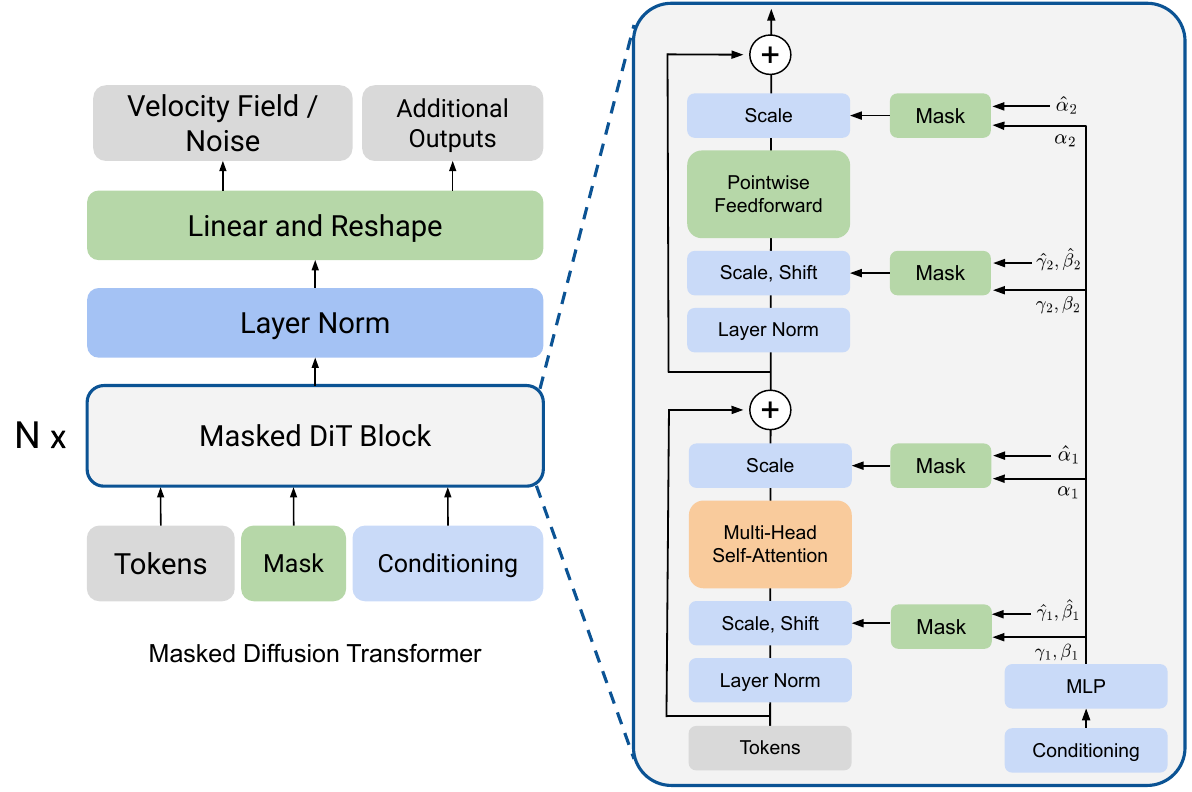}
    \caption{
        \textbf{Masked DiT block.} Each block applies adaptive layer normalization (adaLN) with two independent sets of modulation parameters, one for target (masked) tokens and one for given (unmasked) tokens, both conditioned on the diffusion timestep embedding $t$. The target branch learns denoising-appropriate scale and shift, while the given branch preserves clean conditioning signals with minimal distortion. 
    }
    \label{fig:masked_dit_block}
\end{figure}

Masked DiT blocks resolve this by maintaining two independent adaLN parameter sets within each block. Let $(\gamma^{\text{tgt}}, \beta^{\text{tgt}})$ and $(\gamma^{\text{gvn}}, \beta^{\text{gvn}})$ denote the scale and shift vectors produced by two separate MLPs that both take the diffusion timestep embedding as input. For a token $z_i$ entering a layer norm, the modulated output is
\begin{equation}
    \hat{z}_i = 
    \begin{cases}
        \gamma^{\text{tgt}} \odot \mathrm{LN}(z_i) + \beta^{\text{tgt}}, & \text{if } i \in \mathcal{P} \text{ (target)}, \\[4pt]
        \gamma^{\text{gvn}} \odot \mathrm{LN}(z_i) + \beta^{\text{gvn}}, & \text{if } i \in \mathcal{G} \text{ (given)},
    \end{cases}
    \label{eq:masked_adaln}
\end{equation}
where $\mathcal{P}$ and $\mathcal{G}$ are the sets of target and given token indices respectively, and $\mathrm{LN}$ denotes standard layer normalization. Both MLP heads share the same timestep embedding but are otherwise independent, allowing the network to learn qualitatively different modulation regimes for the two token roles.

This dual-modulation design has several consequences. First, the target branch can learn noise-level-dependent scaling that amplifies or suppresses features as appropriate for the current denoising step, analogous to standard DiT behavior. Second, the given branch can learn near-identity modulation (scale close to one, shift close to zero) that preserves the information content of clean conditioning tokens across layers. Third, because the two branches share the same attention computation after modulation, target tokens still attend to given tokens and vice versa, maintaining the information flow needed for conditional generation. The separation occurs only at the normalization and modulation stage, not at the attention or feedforward stages.

In practice, we apply this dual modulation at both the pre-attention and pre-feedforward layer norms within each DiT block. The attention itself follows the structured spatiotemporal factorization described in the main text, with spatial attention coupling motion and action tokens within a timestep, temporal attention linking the same keypoint across timesteps, and context attention connecting all target tokens to observation and task tokens. The mask that determines which tokens are target versus given is fixed for a given training sample and inference call, so the routing in \Cref{eq:masked_adaln} adds negligible computational overhead beyond the additional MLP parameters for the second modulation head.

\subsection{Training Hyperparameters}
We train \MethodAbbr~with AdamW using a learning rate of $1 \times 10^{-4}$, weight decay of $1 \times 10^{-6}$, batch size of $16$, and gradient clipping with maximum norm $1$. Training runs for $16$ epochs on $8$ NVIDIA A100 GPUs, with a linear warmup over the first $10000$ optimization steps. The pretraining objective uses loss weights $\lambda_x = 1$, $\lambda_a = 2$, and $\lambda_c = 0.001$ for motion reconstruction, action reconstruction, and contrastive task latent learning respectively. The pretraining mixture contains approximately 10M steps of human video data (HOI4D and Xperience), 10M steps of real robot data (DROID), and 10M steps of simulated robot data, sampled with equal probability during training.

The transformer backbone uses $6$ layers with hidden size $768$ and $12$ attention heads. Each layer follows the structured attention design described in the main paper, alternating among spatial attention over same timestep motion and action tokens, temporal attention over the same keypoint across timesteps, and context attention from target tokens to observation and task tokens. The prediction horizon is $H=16$, and the action chunk length is set to the same horizon. For each sample, the number of tracked object keypoints satisfies $K \in [1,16]$.

\section{Data Pipeline}

We detail the data curation pipelines used to construct the pretraining mixture for \MethodAbbr. The real-world pipeline extracts 3D keypoint trajectory supervision from monocular RGB videos across diverse sources, while the simulated pipeline generates randomized manipulation trajectories with full state access.

\subsection{Real-World Data Pipeline}

To extract 3D keypoint trajectory supervision from monocular RGB videos, whether from real robot demonstrations (DROID~\citep{khazatsky2024droid}), human videos (HOI4D~\citep{Liu_2022_hoi4d}, Xperience~\citep{xperience_10m}), or cross-embodiment sources, we apply a four-stage pipeline that requires no ground-truth depth, camera intrinsics, or action labels.

\begin{figure}[t]
    \centering
    \includegraphics[width=0.95\textwidth]{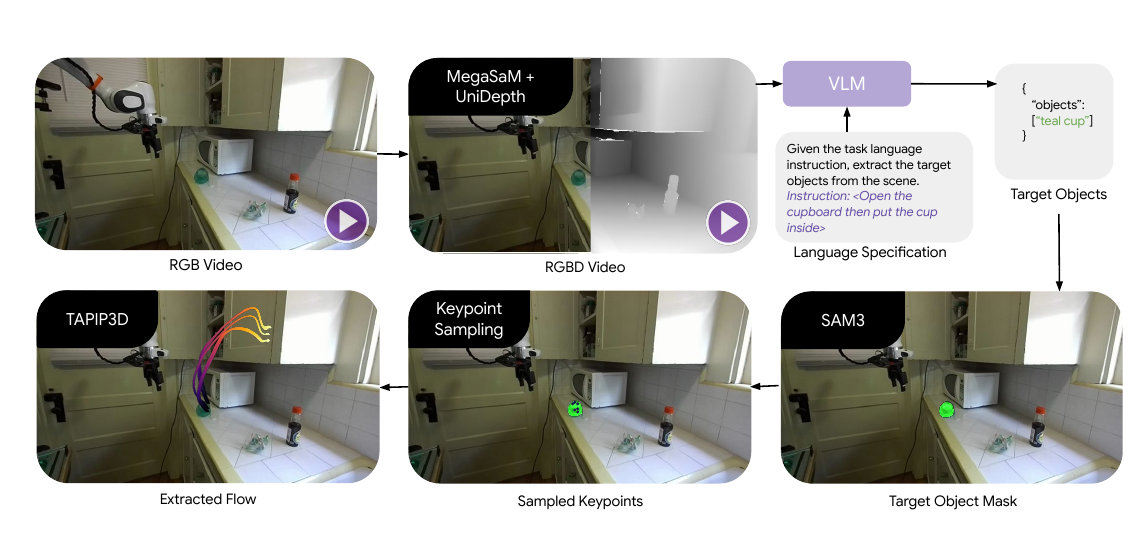}
    \caption{\textbf{Data pipeline.} We extract 3D keypoint trajectory supervision from monocular RGB videos by estimating camera motion and depth, aligning depth to metric scale, segmenting and sampling task-relevant object points, and tracking the resulting 3D keypoints over time.}
    \label{fig:data_pipeline}
\end{figure}

\textit{Stage 1: Camera and depth estimation.} We first run MegaSaM~\citep{li2025megasam} on each video to jointly estimate per-frame camera poses, camera intrinsics, and normalized (up-to-scale) depth maps. MegaSaM recovers consistent structure and motion from dynamic videos, providing a geometrically coherent reconstruction of the scene across frames.

\textit{Stage 2: Metric depth alignment.} The depth maps produced by MegaSaM are scale-ambiguous. To recover metric-scale 3D geometry, we use UniDepth~\citep{piccinelli2024unidepth} to predict a metric depth map for a reference frame, then align the MegaSaM depth sequence to this metric scale via a global scale-and-shift fit. This yields per-frame metric depth maps consistent with the estimated camera trajectory.

\textit{Stage 3: Keypoint sampling.} To focus tracking on task-relevant regions, we identify the object of interest in each video by prompting a vision-language model (GPT-5-nano) with sub-sampled keyframes from the video. For datasets with language annotations, we prompt the vision-language model with the language instruction and a single frame from the video to identify the target object. The language descriptor of the target object is then passed to SAM~3~\citep{carion2025sam} to produce a segmentation mask. We sample keypoints densely on the segmented object surface to ensure fine-grained coverage of the manipulation target. In addition, we randomly sample a smaller set of points elsewhere in the scene to provide background context and allow the model to capture overall scene dynamics.

\textit{Stage 4: 3D point tracking.} Given the metric depth maps, camera intrinsics, camera poses, and sampled query points from the previous stages, we apply TAPIP3D~\citep{tapip3d} to produce persistent 3D keypoint trajectories $\{x_{k,t}\}$ that serve as the object-motion supervision signal described in the main paper.

For data which includes ground-truth depth and camera parameters, Stages 1 and 2 are bypassed and only Stages 3 and 4 are applied. 

\subsection{Simulated Data Pipeline}

We adopt the simulated data generation pipeline from COIL~\citep{coil2025}. The pipeline constructs randomized tabletop manipulation scenes with a Franka Research 3 robot matching our real-world setup. Scenes are populated with diverse assets drawn from the YCB dataset~\citep{calli2015ycb}, PartNet-Mobility~\citep{Xiang_2020_SAPIEN}, Mujoco Scanned Objects~\citep{downs2022scannedobjects,zakka2022scannedobjectsmujoco}, and publicly available 3D models, spanning manipulable objects, tools with annotated functional regions, and containers. 

Two heuristic action primitives generate trajectories. The first grasps a random object and moves it along a B\'{e}zier curve with randomized 6-DoF waypoints. The second performs tool-use interactions (sweeping, poking, hooking) by grasping a tool at its functional region and translating it toward a target object. Only successful trajectories are retained, and motion labels are assigned in hindsight after the rollout is completed.

\section{Additional Details of Experimental Setup}

We provide a detailed account of the experimental setup used to evaluate \MethodAbbr~on real-world tabletop manipulation tasks. In this section, we describe the environment setup, task design, and evaluation protocol to enable reproducibility. 

\subsection{Environment Setup}
We follow the DROID environment setup~\citep{khazatsky2024droid}, which uses a Franka Research FR3 robot arm equipped with a Robotiq 2F-85 parallel-jaw gripper. During action-generation evaluations, the policy sends target end-effector poses and gripper actions at 15Hz. A ZED stereo camera is mounted on each side of the robot to provide visual coverage of the workspace. In our evaluation, only the right camera is used to provide RGBD observations to the policy.

\subsection{Task Design}
We design the real-world evaluation tasks with out-of-distribution objects to test whether the policy generalizes beyond the objects observed during pretraining. The benchmark contains three tabletop manipulation tasks.

\textbf{Insertion} requires the robot to pick up a utensil from the table and place it fully into a utensil holder. This task demands precise 6-DoF positioning to align the utensil with a narrow opening, making it sensitive to small orientation errors during both the grasp and the placement phase.

\textbf{Sweeping} requires the robot to grasp a brush and sweep paper debris into a dustpan. This is a tool-use task in which the robot must coordinate the brush motion relative to both the debris and the dustpan, requiring accurate contact-rich control through an intermediary object rather than direct grasping.

\textbf{Folding} requires the robot to fold a pair of blue jeans from one side of the table to the other. Manipulating deformable materials is challenging because the object's shape changes continuously during execution, requiring the policy to reason about non-rigid geometry and produce smooth, extended motions that achieve the target fold.

Across trials, we randomize object poses while keeping the task semantics fixed. Each task is evaluated over 20 trials.

\subsection{Evaluation Protocol}
\textbf{Success criteria. } We use binary success labels for all real-world trials. An Insertion trial is successful if the utensil lies completely inside the utensil holder at the end of the episode. A Sweeping trial is successful if the paper debris is contained inside the dustpan. A Folding trial is successful if the final folding angle of the jeans is less than $10^\circ$ from the target fold.

\textbf{Reference motion specification. } For motion-conditioned evaluation, a human provides a sparse reference motion through an interactive UI. We use one reference motion per trial. The reference keypoints are sampled randomly on the surface of the object of interest, and the human specifies their desired motion to define the task intent for that trial.

\textbf{Few-shot adaptation.} For few-shot adaptation, demonstrations are collected in the same test environment as evaluation. In the action-supervised setting, the demonstrations are collected through teleoperation and provide paired observation-action supervision. In the motion-supervised setting, the demonstrations are human videos recorded from the same scene and viewpoint, providing object-motion supervision without robot action labels.

\textbf{Dynamics prediction evaluation.} For dynamics prediction, we evaluate object-motion forecasts over a horizon of 10 future steps using 16 keypoints per scene. Ground-truth 3D keypoint trajectories are obtained with TAPIP3D~\citep{tapip3d}. We report mean squared error averaged over 3D keypoint coordinates, keypoints, timesteps, tasks, and episodes.

\section{Multimodal Task Conditioning}

\begin{figure}[t]
    \centering
    \includegraphics[width=1.0\textwidth]{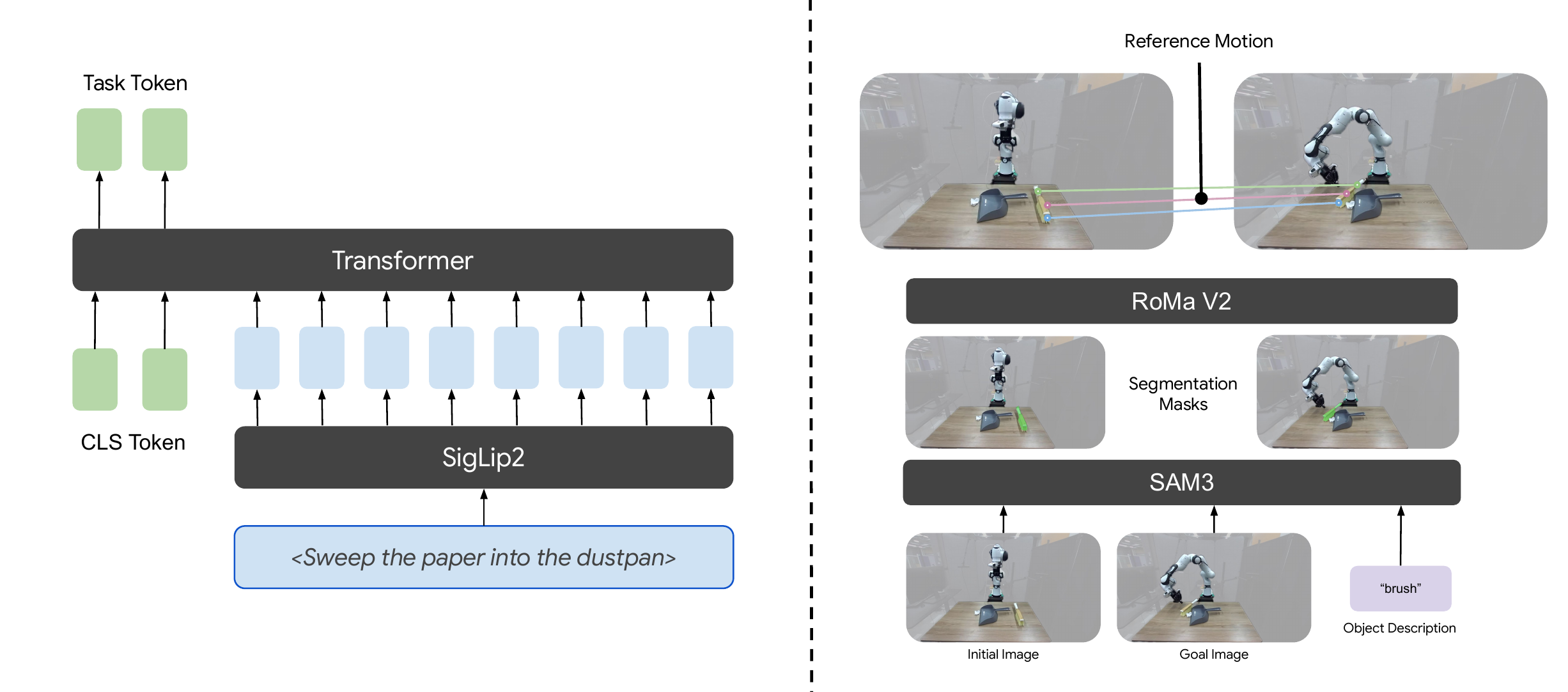}
    \caption{
        \textbf{Multimodal task conditioning.} Left: instruction following replaces the motion-derived task latent with tokens from a text encoder. Right: goal reaching uses a user-provided object description, SAM~3 segmentation, and RoMaV2-based 2D point matching to convert a goal image into a sparse start-to-end reference motion.
    }
    \label{fig:additional_inference}
\end{figure}

\begin{figure*}[t]
    \centering
    \includegraphics[width=0.95\textwidth]{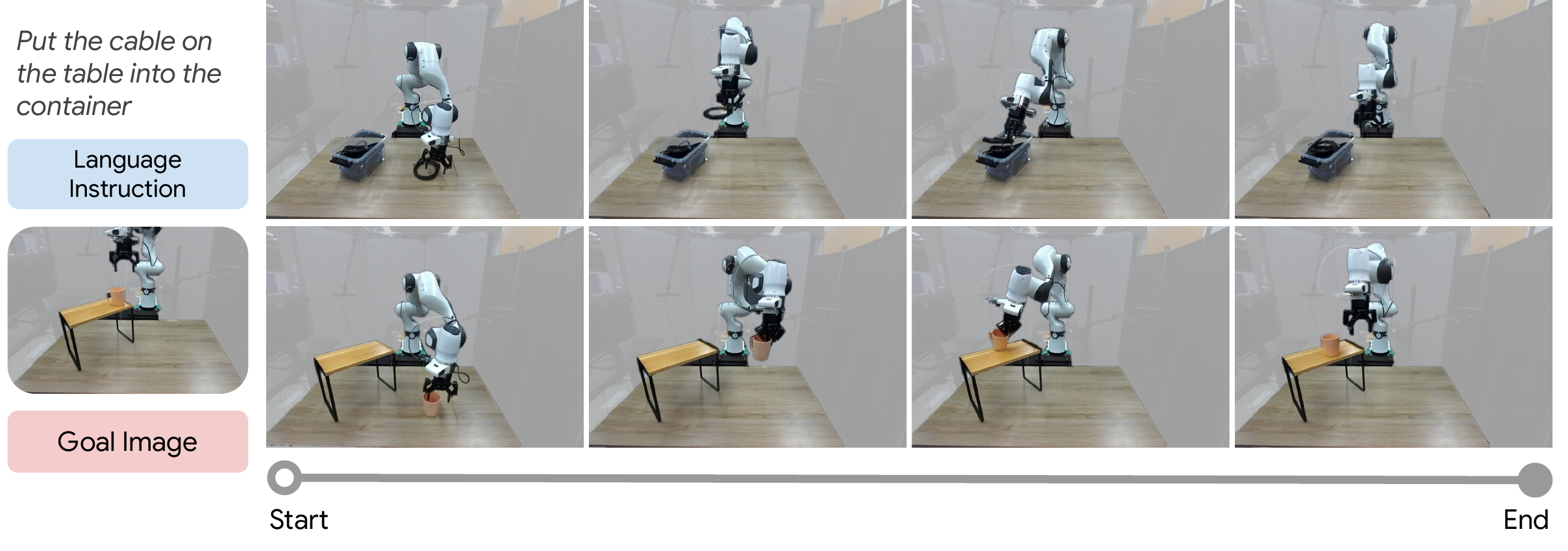}
    \caption{
        \textbf{Task execution under alternative inference modes.} The same pretrained \MethodAbbr~checkpoint performs instruction following (top rows) and goal reaching (bottom rows) without retraining. In instruction-following mode, a text instruction replaces the reference motion and the language encoder produces the task latent. In goal-reaching mode, a goal image is converted into a sparse two-timestep reference motion via the keypoint extraction pipeline described in \Cref{fig:additional_inference}, which then feeds the standard motion-conditioned pathway. Across both modes, the robot produces coordinated manipulation behavior that respects the expressed task intent.
    }
    \label{fig:inference_task_exec}
\end{figure*}

The masked generative formulation of \MethodAbbr~allows the same pretrained model to be conditioned on alternative task specifications beyond reference motion by replacing the task latent $c$. As shown in \Cref{fig:additional_inference}, we present two examples of such modes: instruction following, where $c$ is derived from a text instruction, and goal reaching, where $c$ is derived from a goal image. Both require no retraining of the denoising transformer. \Cref{fig:inference_task_exec} provides qualitative rollouts showing that the same pretrained checkpoint produces meaningful manipulation behavior under these alternative conditioning modes.

\subsection{Instruction Following}

The main paper introduces the task encoder $h_\psi(o_0, x_c)$, which encodes a reference motion $x_c$ together with the initial observation $o_0$ into $M$ task tokens comprising $c$. Because the contrastive objective $\mathcal{L}_c$ trains $c$ to capture task intent independently of the specific modality through which the intent is specified, $c$ can also be produced by a separate language encoder that shares the same latent space. Concretely, a language encoder $h_\omega$ processes a text instruction $\ell$ in two stages. First, a \VisualEncoder~text encoder tokenizes $\ell$ into a sequence of 768-dimensional embeddings with a maximum length of 64. $M$ learnable classification tokens are prepended to the sequence, and the resulting tokens are passed through $N_\ell$ transformer layers that alternate self-attention among the language tokens with cross-attention from a set of $M$ learnable aggregation queries. The output is $M$ task tokens that replace $c$ at inference time. This architecture mirrors the aggregation stage of the keypoint specification encoder $h_\psi$, ensuring that both pathways produce tokens in the same latent space and enabling the contrastive objective to align them.

At inference time, instruction following proceeds exactly like motion-conditioned control, except that $c = h_\omega(\ell)$ replaces $c = h_\psi(o_0, x^\star)$. The denoising transformer, observation encoder, and all other parameters remain unchanged. 

\subsection{Goal Reaching}

goal reaching specifies the task through a goal image $o_g$ depicting the desired final configuration of the scene, together with a user-provided object description, rather than through a reference motion or language instruction. Because \MethodAbbr~requires a reference motion as input to $h_\psi$, goal reaching converts the goal image into a sparse two-timestep reference motion via an automated keypoint extraction pipeline and then proceeds with standard motion-conditioned inference.

Given the initial observation image $o_0^{\text{rgb}}$, the goal image $o_g^{\text{rgb}}$, and their corresponding depth maps $o_0^{\text{depth}}$ and $o_g^{\text{depth}}$, the pipeline operates in four stages. First, the user provides an object description, and SAM~3~\citep{carion2025sam} segments the described object in both images. Second, RoMaV2~\citep{edstedt2025romav2} predicts dense 2D point correspondences between the masked regions in the two images. Third, each matched 2D keypoint pair is lifted to 3D camera coordinates using the corresponding depth map and calibrated camera intrinsics via bilinear depth interpolation and inverse projection. Keypoints with invalid or zero depth are marked as invalid. Fourth, the resulting 3D keypoint pairs are assembled into a reference motion $x_c$ with $T = 2$ timesteps and up to $K = 16$ keypoints, where timestep 0 contains the 3D positions in the initial scene and timestep 1 contains the corresponding 3D positions in the goal scene. This reference motion is encoded through the standard keypoint specification encoder $h_\psi(o_0, x_c)$ and fed to the denoising transformer exactly as in the motion-conditioned control mode.

Because the extracted keypoints provide only start and end positions rather than a dense trajectory, the contrastive training of $h_\psi$ is essential for making the task latent $c$ invariant to the sparsity and ordering of the reference motion. The same SE(3) augmentation and keypoint subsampling applied during pretraining ensure that $h_\psi$ generalizes from dense multi-timestep reference motions used during training to the sparse two-timestep references produced by goal images at deployment. No fine-tuning or retraining is needed.

\section{Scalability Analysis}

\begin{figure}[t]
    \centering
    \includegraphics[width=0.5\textwidth]{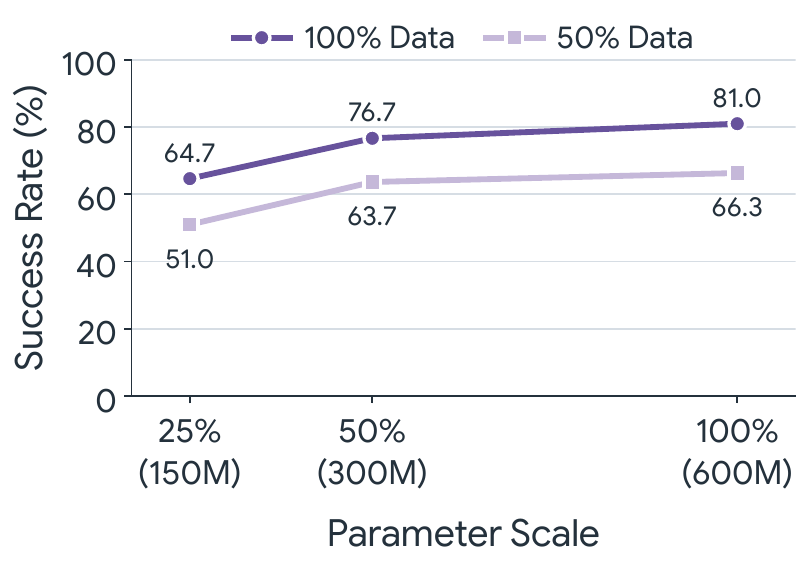}
    \caption{
        \textbf{Data and model scaling analysis.} We report the average success rate across the three simulation tasks for six configurations shown as line charts. Both data scale and model parameter scale contribute to performance, with full model achieving the strongest result.
    }
    \label{fig:scaling_analysis}
\end{figure}

We evaluate the scalability of \MethodAbbr~by varying both the amount of training data and the model size. \Cref{fig:scaling_analysis} reports the average success rate over the simulated tasks, each evaluated with 100 trials. These results suggest that both data scale and parameter scale contribute to performance, with the full-data 600M-parameter model achieving the strongest result.

\section{Additional Qualitative Results}
We provide additional qualitative results to show how the same pretrained \MethodAbbr~checkpoint executes tasks with different manipulation structure. The examples cover rigid object insertion, tool mediated sweeping, and deformable folding, matching the real world tasks used in the quantitative evaluation. These rollouts are intended to illustrate the breadth of behaviors induced by the shared motion action interface rather than introduce a separate evaluation protocol.

\begin{figure*}[t]
    \includegraphics[width=0.95\textwidth]{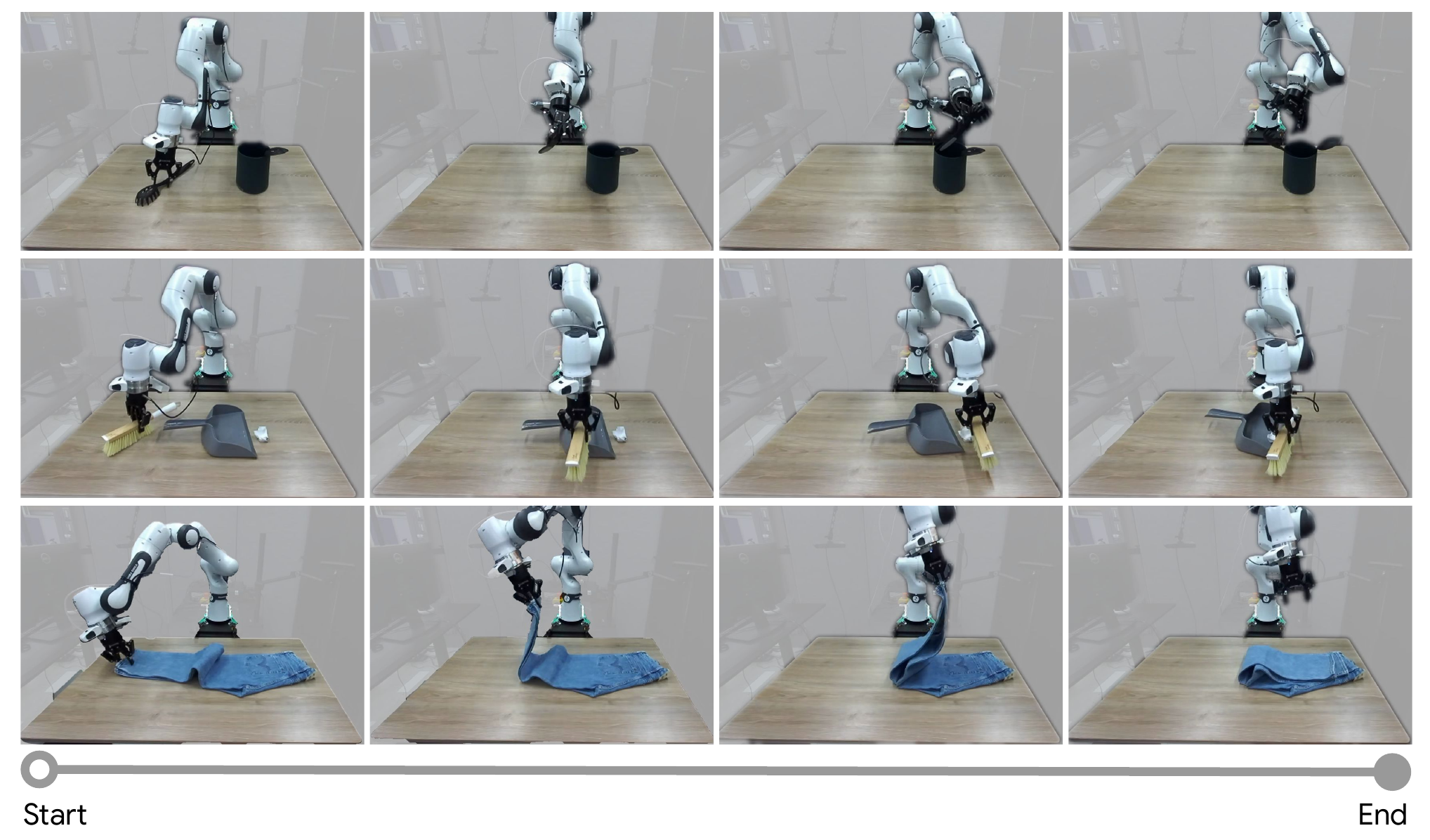}

    \caption{
        \textbf{Task execution.} We show representative rollouts of \MethodAbbr~on the three real world evaluation tasks. The same pretrained model executes rigid object insertion, tool use, and deformable folding by conditioning on task motion and replanning from the current observation.
    }
    \label{fig:task_execution}
\end{figure*}

As shown in \Cref{fig:task_execution}, during closed loop executions, \MethodAbbr~conditions on the current observation and a task latent derived from reference motion, predicts an action chunk over the same horizon used during training, and replans after executing the chunk. The resulting behavior preserves the task level object motion across diverse interaction modes, while allowing the robot trajectory to adapt to the current object pose and scene geometry.

%===============================================================================

\end{document}